\documentclass{article}

\PassOptionsToPackage{numbers, compress}{natbib}

\usepackage[preprint]{neurips_2023}

\usepackage[utf8]{inputenc} %
\usepackage[T1]{fontenc}    %
\usepackage{hyperref}       %
\usepackage{url}            %
\usepackage{booktabs}       %
\usepackage{amsfonts}       %
\usepackage{nicefrac}       %
\usepackage{microtype}      %
\usepackage[dvipsnames]{xcolor}         %
\usepackage{graphicx}
\usepackage{subcaption}
\usepackage{caption}
\usepackage{algorithm,algpseudocode}
\usepackage{tabularx}
\usepackage{enumitem}
\usepackage{wrapfig}

\usepackage{hyperref}

\usepackage{amsmath}
\usepackage{amssymb}
\usepackage{mathtools}
\usepackage{amsthm}

\usepackage[capitalize,noabbrev]{cleveref}

\theoremstyle{plain}

\theoremstyle{definition}

\theoremstyle{remark}

\usepackage[textsize=tiny]{todonotes}

\title{On the Efficacy of 3D Point Cloud Reinforcement Learning}

\author{%
  Zhan Ling\thanks{Equal contribution}\quad Yunchao Yao\footnotemark[1]\quad Xuanlin Li\quad Hao Su\\
  UC San Diego
}

\begin{document}

\maketitle

\begin{abstract}
Recent studies on visual reinforcement learning (visual RL) have explored the use of 3D visual representations. However, none of these work has systematically compared the efficacy of 3D representations with 2D representations across different tasks, nor have they analyzed 3D representations from the perspective of agent-object / object-object relationship reasoning. In this work, we seek answers to the question of when and how do 3D neural networks that learn features in the 3D-native space provide a beneficial inductive bias for visual RL. We specifically focus on 3D point clouds, one of the most common forms of 3D representations. We systematically investigate design choices for 3D point cloud RL, leading to the development of a robust algorithm for various robotic manipulation and control tasks. Furthermore, through comparisons between 2D image vs 3D point cloud RL methods on both minimalist synthetic tasks and complex robotic manipulation tasks, we find that 3D point cloud RL can significantly outperform the 2D counterpart when agent-object / object-object relationship encoding is a key factor. Our code will be released at \url{https://github.com/lz1oceani/pointcloud_rl}.
\end{abstract}

\section{Introduction}

Building sample-efficient agents that solve challenging tasks given visual observations is a long-standing problem in reinforcement learning (visual RL). In recent years, visual RL has achieved significant progress across diverse tasks such as control \citep{hafner2019dream, parisi2022unsurprising} and robotic manipulation \citep{Zhu2018ReinforcementAI, Dasari2019RoboNetLM,Chen2021ASF}. For visual RL agents to accomplish many real world tasks, it is often beneficial if agents are capable of reasoning the spatial \textbf{\emph{relationship}} between themselves and the surrounding objects as well as among surrounding objects. For example, when a robot arm learns to open a cabinet door, if the agent already knows about the relative direction between the robot hand and the door handle, it can learn to simply move its hand towards the direction of the door handle, thereby reducing the learning sample complexity.

Much prior work on visual RL has adopted RGB or RGB-D images as visual input. However, objects in the real world live in 3D space. When we move the camera away or closer to an object, the number of pixels an object occupies on the 2D image can vary greatly, leading to potential perception difficulties. In addition, neighboring pixels might have distinct depth and encode distinct spatial locations, making it challenging for 2D CNNs to reason about and precisely localize objects in 3D. On the other hand, for visual representations native to the 3D world and have explicit coordinate information in $\mathbb{R}^3$, such as point clouds and 3D volumes, the object size is invariant to camera positioning, and neighboring points / volumes entail spatial proximity. Furthermore, recent years have witnessed remarkable progress in building neural networks for 3D understanding~\citep{qi2017pointnet,qi2017pointnet++,Thomas2019KPConvFA,Zhao2020PointT}.  For applications that involve agent-object and object-object relationship reasoning, such as robot grasping~\citep{Sundermeyer2021ContactGraspNetE6,Shridhar2022PerceiverActorAM}, 6D pose estimation~\citep{wang2019densefusion,he2020pvn3d}, and agent-object/object-object interaction affordance prediction~\citep{mo2021o2oafford,wang2022adaafford}, we are witnessing the trend to move from understanding 2D images (including RGB-D data) to 3D native representations. 
Therefore, it is worth exploring whether 3D visual representations like point clouds, along with their inductive biases, could benefit visual RL. Theoretically, to benefit RL training, we would like the policy/value networks to generalize to unseen states. \emph{Naturally, the inductive bias of 2D networks tends to generate features that respect the 2D input similarity, while 3D networks generate features that respect 3D input similarity. This difference makes 3D networks potentially better suited for generalizing policy/value networks on tasks that heavily rely on 3D spatial relationships.}

While there have been very recent works on visual RL from 3D representations like point clouds and 3D volumes \citep{huang2021generalization,liu2022frame,coarsetofineqattention,zhou2023learning}, none of these works has systematically compared the efficacy of 3D vs. 2D representations in RL across different tasks, nor did they analyze 3D representations from the agent-object/object-object relationship reasoning perspectives. Unfortunately, the field currently lacks a comprehensively-evaluated RL algorithm with 3D input, and RL researchers do not generally believe in the necessity of using 3D neural networks which learn representations natively in 3D space for policy/value functions. This raises an important question for representation learning in RL: \textit{When and how do 3D representations, such as point clouds, provide a beneficial inductive bias for RL?}

 As an early attempt to explore the gap between 3D \& 2D visual RL, in this work, we make the following contributions: \textbf{(1)} We systematically study and compare visual RL using 2D images vs. 3D representations, specifically 3D point clouds, across different robotic manipulation and control tasks. \textbf{(2)} We carefully investigate the best practices and design choices for 3D visual RL from point clouds, such as observation post-processing, data augmentation, and network architectures. We hope that our study provides insights and guidance for future works on 3D visual RL. \textbf{(3)} We find that for many tasks that do not involve agent-object or object-object relationship reasoning (e.g., many tasks in DMControl~\citep{tassa2018deepmind}), the performance of 3D point cloud RL is close to that of 2D visual RL. On the other hand, for robotic manipulation tasks that involve heavy spatial relationship reasoning, we find that 3D point cloud RL achieves better sample efficiency and performance. We further analyze how 3D point cloud representations could benefit RL by having better inductive bias on encoding spatial relationships using minimal synthetic experiments and complex robotic manipulation experiments.

\begin{figure*}[t!]
    \begin{minipage}[b]{0.51\linewidth}
        \label{teaser}
        \includegraphics[width=\linewidth]{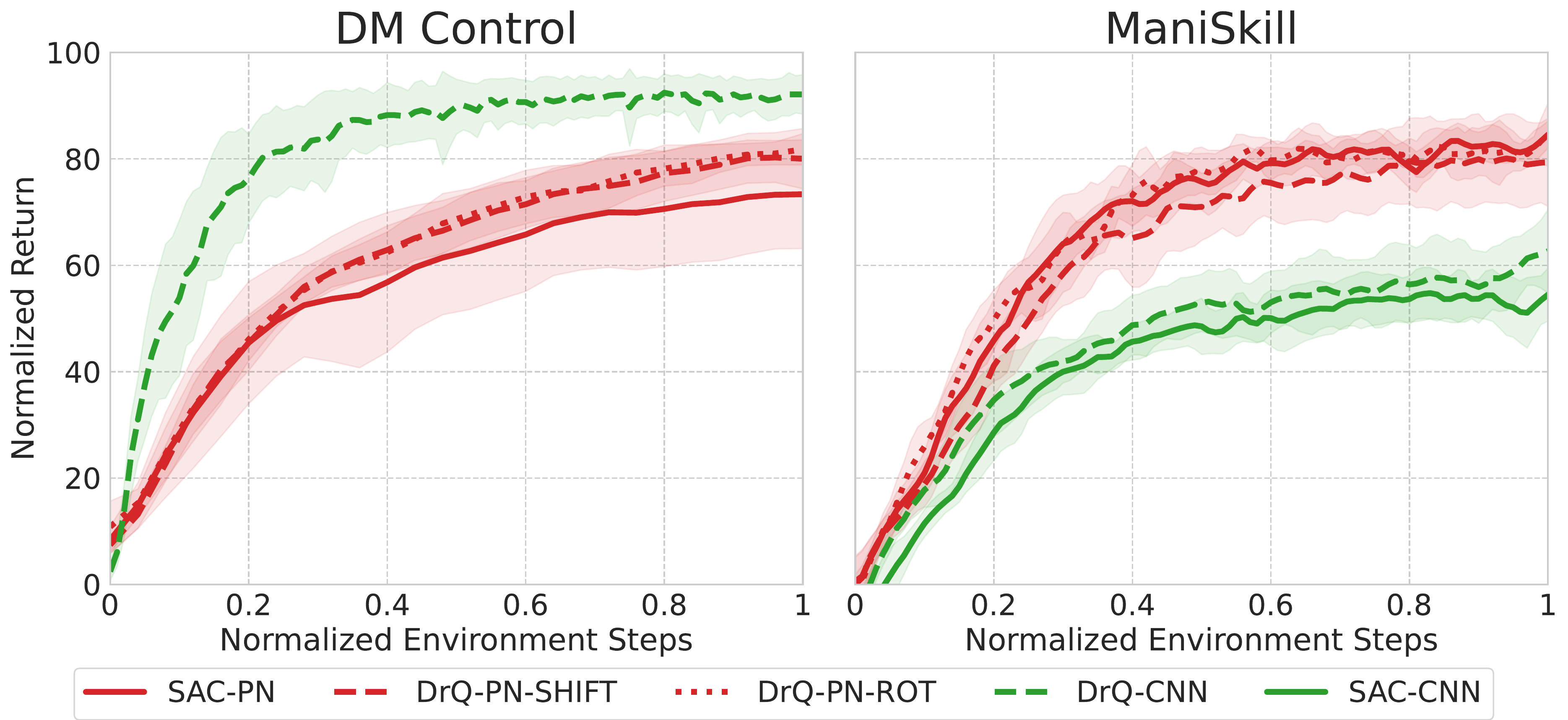}
        \subcaption{Training curves for DM Control and ManiSkill tasks.}
        \label{fig:teaser}
    \end{minipage}
    \begin{minipage}[b]{0.48\linewidth}
    \begin{minipage}[b]{0.55\linewidth}
        \includegraphics[width=\linewidth]{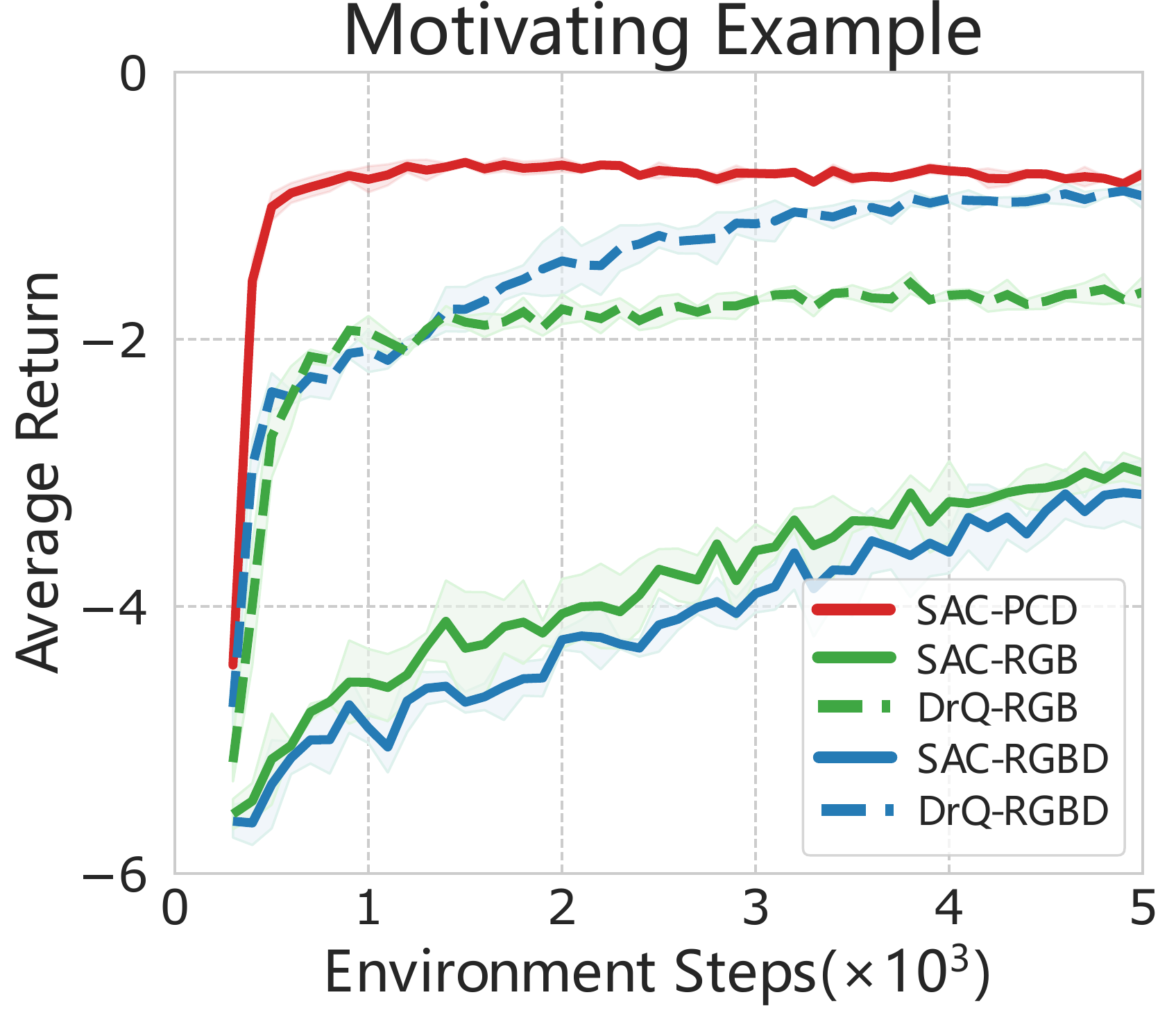}
    \end{minipage}
    \begin{minipage}[b]{0.40\linewidth}
        \centering
        \begin{minipage}[b]{\linewidth}
        \centering
        \begin{minipage}[b]{0.385\linewidth}
        \includegraphics[width=\linewidth]{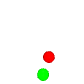}
        \end{minipage}
        \begin{minipage}[b]{0.385\linewidth}
        \includegraphics[width=\linewidth]{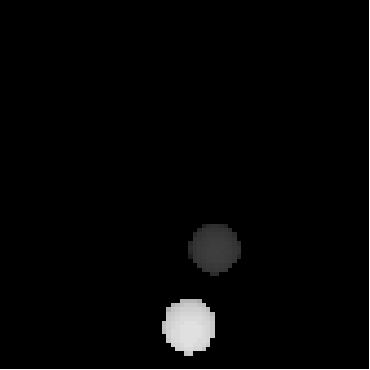}
        \end{minipage}
        \small{RGB \& Depth Image}
        \vspace{0.2em}
        \end{minipage}\\
        \begin{minipage}[b]{\linewidth}
        \centering
        \begin{minipage}[b]{\linewidth}
        \includegraphics[width=0.76\linewidth]{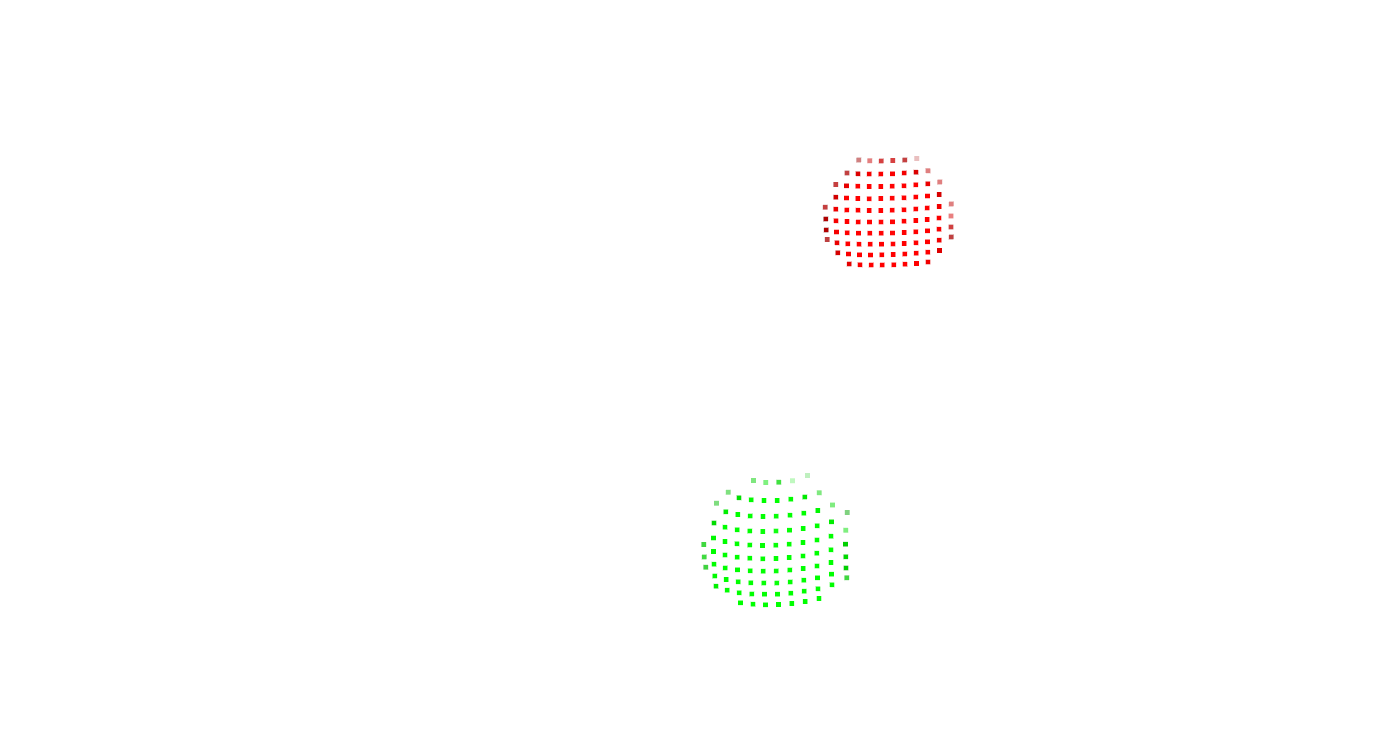}
        \end{minipage}
        \small{Point Cloud}
        \end{minipage}
    \end{minipage}
    \subcaption{Our minimalist motivating example.}
    \label{fig:illustrative_env}
    \end{minipage}
    \caption{\textbf{a)} \textbf{(Left)} Normalized return on 6 continuous control tasks from the PlaNet Benchmark~\citep{laskin2020reinforcement}: \textit{Ball-in-cup Catch}, \textit{Cartpole Swingup}, \textit{Cheetah Run},
    \textit{Finger Spin}, \textit{Reacher Easy}, and \textit{Walker Walk}. These tasks are mostly 2D locomotion tasks that involve \textit{little agent-object and object-object relationship reasoning}, and 2D visual RL agents perform the best. \textbf{(Right)} Normalized return on 4 robot manipulation tasks from the ManiSkill Benchmark~\citep{mu2021maniskill}: \textit{OpenCabinetDrawer}, \textit{OpenCabinetDoor}, \textit{PushChair}, and \textit{MoveBucket}. These tasks involve \textit{heavy robot-object relationship reasoning in 3D scenes}, where 3D point clouds could benefit RL by having better inductive bias on encoding such spatial relationships. (We only include RGB agents here as RGB-D agents perform similarly). \textbf{b)} Illustration of our minimalist motivating example for 3D visual RL from point clouds (environment built in MuJoCo~\citep{todorov2012mujoco}). An agent needs to move from the red sphere to the green sphere (sphere sizes are randomized). 3D point cloud RL agents achieve the best sample efficiency and final performance (though agents do not perfectly solve the task since occlusions can occur during initialization).
    }
\end{figure*}
\section{Related Work}

\textbf{Visual Reinforcement Learning.} Vision-based reinforcement learning has achieved tremendous progress in applications such as robotic manipulation \citep{Kalashnikov2018QTOptSD, Zeng2020TransporterNR, Jang2022BCZZT, Nair2022R3MAU}, control \citep{hafner2019dream, parisi2022unsurprising, xiao2022masked, hansen2022temporal}, and playing games \citep{mnih2013playing, berner2019dota, ye2020mastering, reed2022generalist}. To improve the sample efficiency and the generalizability of visual RL agents, prior work has adopted different lines of approaches, such as domain randomization \citep{Tobin2017DomainRF,Peng2017SimtoRealTO,Pinto2017AsymmetricAC,Mehta2019ActiveDR}, self-supervised representation learning \citep{Yarats2019ImprovingSE,srinivas2020curl,hafner2020mastering,Sekar2020PlanningTE,xiao2022masked}, and data augmentation \citep{laskin2020reinforcement,kostrikov2020image,yarats2021mastering,hansen2021stabilizing}. However, most prior work adopt RGB or RGB-D images as visual input, and has rarely studied 3D representations such as point clouds. %

\textbf{Reinforcement Learning from 3D Point Clouds.} 
While much prior work has demonstrated that 3D point clouds are powerful visual representations in fields such as perception~\citep{He2021FFB6DAF,Vu2022SoftGroupF3}, self-driving~\citep{Wang2018PseudoLiDARFV,you2022learning} and robotic manipulation~\citep{Simeonov2021NeuralDF,Eisner2022FlowBot3DL3}, it is not until very recently that point clouds start to be used as input visual representations for reinforcement learning \citep{huang2021generalization,Ze2022VisualRL,chen2022system,wu2022learning,liu2022frame}. While these works demonstrate that 3D point cloud can be used for visual RL, their experiments are ablations over their method designs without comparison to 2D visual RL methods that are still the mainstream. We are the first to carefully analyze the efficacy of 3D representations over 2D representations across diverse tasks. Along the process, we carefully investigate the design choices and best practices for 3D visual RL from point clouds, so that they perform on the same stage or significantly better than the 2D visual RL counterparts, depending on the nature of the tasks.

\section{Background}
\label{sec:background}
\textbf{Reinforcement Learning from 2D \& 3D Vision.}
We formulate a visual reinforcement learning (visual RL) task as an infinite-horizon Partially Observable Markov Decision Process (POMDP)~\cite{smallwood1973optimal} represented by the tuple $M = \left(S, A, \mu, T, R, \gamma, \Omega, O\right)$, Here $S, A, \Omega$ are the state space, the action space, and the observation space of an environment; $\mu(s), T(s'|s, a), R(s, a)$, and $\gamma$ are the initial state distribution, the state transition probability, the reward function, and the discounted factor; $O(s)$ is the observation function that maps an environment state $s$ to its corresponding observation in $\Omega$. For visual RL tasks studied in this work, $O(s)$ consists of $K (K\ge 1)$ consecutive frames of 2D images / 3D point clouds. If $O(s)$ consists of RGB / RGB-D images, then $O(s) \in \mathbb{R}^{K\times H\times W \times C}$, where $C = 3$ for RGB and $C = 4$ for RGB-D. If $O(s)$ consists of point clouds, then $O(s) \in \mathbb{R}^{K\times N \times C}$, where $N$ is the number of points in each frame of point cloud; $C=3+C' (C' \ge 0)$ is the point cloud feature dimension, which consists of 3-dimensional point coordinates along with optional features such as RGB colors. The agent $\pi:\Omega \to A$ aims to optimize the expected accumulated returns from the environment by maximizing $J_\pi = \mathbb{E}_{\mu, \pi, T}\left[\sum_{t=0}^{\infty} \gamma^t R(s_t, a_t)\right]$.

\textbf{Soft Actor-Critic~(SAC)~\cite{haarnoja2018soft}} is a maximum-entropy off-policy actor-critic algorithm for continuous control tasks which iteratively updates the state-action value function $Q_\theta(s, a)$ and the stochastic policy actor $\pi_\phi(a|s)$, where $Q_\theta$ is trained to minimize the soft Bellman objective $\mathbb{E}_{s_t,a_t}[r(s_t,a_t) + \gamma \mathbb{E}_{s_{t+1},a_{t+1}\sim \pi_{\phi}(\cdot|s_{t+1})}(\bar{Q}_{\theta}(s_{t+1},a_{t+1}) - \alpha \log \pi_{\phi}(a_{t+1} | s_{t+1})) - Q(s_t,a_t)]^2$, and $\pi_{\phi}$ is trained to minimize $D_{KL}[\pi_{\phi}(\cdot | s) || \exp (\frac{1}{\alpha}Q(s, \cdot))]$. Here $\bar{Q}$ is the target $Q$ network, and $\alpha > 0$ is the temperature parameter that trades between exploration vs. exploitation.

\textbf{Data-Regularized Q (DrQ).}
In 2D image-based RL, data augmentation techniques have proven highly effective for improving agent learning performance and sample efficiency. A prominent work in this direction is DrQ \cite{kostrikov2020image,yarats2021mastering}, which extends actor-critic algorithms by averaging the Q-target and the Q value calculations over $K$ and $M$ augmentations of input images, respectively. In practice, DrQ uses random-shift data augmentation and sets $K=2$ and $M=2$. We adopt their official implementation throughout this work.

\section{Methodology}
In this section, we aim to investigate the benefits of using 3D representations in visual reinforcement learning (RL). Specifically, we focus on 3D point clouds, as they are one of the most common forms of 3D representations. To do this, we conduct a minimalist synthetic experiment that compares the performance of RL in 3D-native point cloud space versus 2D image space (Sec.~\ref{sec:minimalist}). This experiment clearly illustrates the core differences in the inductive biases of 3D versus 2D networks, without the need for extensive parameter tuning. However, since 2D visual RL is much more established than 3D visual RL, to make a fair comparison on complex tasks, we must address the challenge that there currently lacks a well-performing and comprehensively-evaluated 3D visual RL algorithm. We therefore discuss possible key design factors for building a strong 3D point cloud RL algorithm (Sec.~\ref{sec:design_space}), and leave extensive experimental exploration to Sec.~\ref{sec:results}. 

\subsection{Minimalist Motivating Example for 3D Visual RL from Point Clouds}
\label{sec:minimalist}

In this section, we design a minimal synthetic experiment to shed light on the inductive biases of 3D point clouds that facilitate 3D spatial relationship understanding in visual RL. Consider the following continuous bandit environment, \textit{3D reacher}, illustrated in Fig.~\ref{fig:illustrative_env}.
The agent is a randomly-sized red sphere initialized at $x_0 \in [-10,10]^3$, which needs to move to a random target position $x_1 \in [-10,10]^3$ (represented as a randomly-sized green sphere) within a single time step. The action $\Delta x \in \mathbb{R}^3$ encodes the movement of the agent, and the episode return equals $-\|x_1 - (x_0 + \Delta x)\|_2$. During environment initialization, we ensure that $\|x_0 - x_1\|_2 \in [d_{min}, d_{max}]$, i.e., the agent and the target are in proper distance. This environment can be seen as the prototype of a wide range of robot manipulation tasks, e.g., moving the robot hand towards a target object. Success on this environment heavily relies on the accurate 3D spatial relationship reasoning between the two spheres.

In Fig.~\ref{fig:illustrative_env} - left, we compare a 3D point cloud agent trained with SAC, along with RGB image and RGB-D image agents trained with SAC and DrQ. We observe that \textbf{(i)} 3D point cloud-based agents achieve better performance and sample efficiency than RGB \& RGB-D agents. Note that while RGB-D images and 3D point clouds essentially encode the same information (point clouds are back-projected from RGB-D images using the known camera matrices), the \textit{explicit 3D coordinate information} in point clouds allows agents to easily reason the \textit{spatial relationship} between the two spheres by simply subtracting one sphere's coordinate from the other. On the other hand, RGB-D images do not provide such inductive bias, making spatial reasoning more challenging. \textbf{(ii)} RGB-based agents struggle at this task as they suffer from the inherent ambiguity of RGB images, where a smaller sphere closer to the camera could have the same pixel projection as a larger sphere farther from the camera, thereby preventing the agent from learning correct actions.

To sum up, our minimal motivating example illustrates the core difference between the inductive biases of 3D and 2D visual representations, along with how 3D point clouds could significantly benefit agents in tasks that involve heavy spatial reasoning.

\subsection{Design Space of 3D Visual RL from Point Clouds}
\label{sec:design_space}
\begin{algorithm}[t]
  \caption{A Framework for RL from 3D Point Clouds}
  \begin{algorithmic}[1]
    \Require $\pi_\phi$, $Q_\theta$: Parametrized policy and Q networks. $T$: total training time steps. \textcolor{Mahogany}{$f(o)$: point cloud post-processing function, such as downsampling}. \textcolor{ForestGreen}{$\mathrm{stack}(o_{t-k}, \cdots, o_t)$: point cloud frame stacking function.} \textcolor{RoyalBlue}{$\mathrm{aug}(o)$: point cloud augmentation function, e.g. random shift and rotation.}
    \State $\mathcal{D}=\{\}$ \algorithmiccomment{Initialize the replay buffer}
    \State $o_1 = \textrm{Env.Reset()}$
    \State \textcolor{Mahogany}{$o'_1 = f(o_1)$} \algorithmiccomment{Post-process the first point cloud}
    \State $o''_1 = o'_1$
    \For{each timestep $t=1..T$}
    \State $a_{t} \sim \pi_\phi(\cdot|o'_t)$ \algorithmiccomment{Sample next action from policy}
    \State $o_{t+1}, r_t = \textrm{Env.Step}(a_t)$ \algorithmiccomment{Step the environment to obtain new point cloud observation and reward}
    \State \textcolor{Mahogany}{$o'_{t+1}=f(o_{t+1})$} \algorithmiccomment{Post-process the next point cloud}.
    \State \textcolor{ForestGreen}{$o''_{t+1}=\mathrm{stack}(o'_{t-k}, \cdots, o'_t)$} \algorithmiccomment{Run point stacking}.
    \State $\mathcal{D} \leftarrow \mathcal{D} \cup (o''_t, a_t, o''_{t+1}, r_t)$ 
    \algorithmiccomment{Add a transition to the replay buffer}
    \If {using data augmentation}
    \State \textsc{Update}($\mathcal{D},\pi_\phi, Q_\theta, \textcolor{RoyalBlue}{\mathrm{aug}}$) \algorithmiccomment{Update the agent with data-augmented RL, e.g., DrQ.}
    \Else
    \State \textsc{Update}($\mathcal{D},\pi_\phi, Q_\theta$) \algorithmiccomment{Update the agent with RL, e.g., SAC.}
    \EndIf
    \EndFor
  \end{algorithmic}
  \label{alg:3d-rl}
\end{algorithm}
As the field currently lacks a robust algorithm for 3D visual RL, we aim to build one in this work. Specifically, we focus on \textbf{RL from 3D point clouds}, as point clouds are one of the most common forms of 3D representations. In this section, we introduce potential components and choices for algorithmic design, including point cloud visual backbones, point cloud post-processing, point frame stacking, and point cloud data augmentations. We will carefully study these design choices in Sec.~\ref{sec:results}. We describe our general framework for 3D point cloud RL in Algorithm \ref{alg:3d-rl}.

\textbf{Point cloud backbone architectures.} In recent years, 3D point cloud deep learning has flourished, with many network architectures being designed for diverse downstream applications such as perception, robotics, and self-driving. When we train an online visual RL agent, we desire the network to be lightweight and capable of fast inference, since RL tasks typically require at least hundreds of thousands of online samples to solve, and we would like to minimize the overhead imposed by the backbone visual network. In this work, we study two 3D commonly-used point cloud architectures: PointNet~\citep{qi2017pointnet} and 3D SparseConvNet~\citep{tang2020searching}. PointNet is a famous light-weight architecture for 3D understanding. It projects point features through a series of light MLPs, then maxpools over the number-of-points dimension to obtain a global point cloud feature. 3D SparseConvNet is a sparse convolution-based architecture for point cloud understanding that achieves rapid inference speed. Compared to PointNet, 3D SparseConvNet has a closer inductive bias to 2D CNNs. Later, we will investigate how such inductive bias difference among 3D architectures impact agent performance.

\begin{figure*}[t!]
 \begin{minipage}{\linewidth}
    \begin{minipage}[b]{0.59\linewidth}
        \begin{minipage}[b]{\linewidth}
            \includegraphics[width=0.325\linewidth]{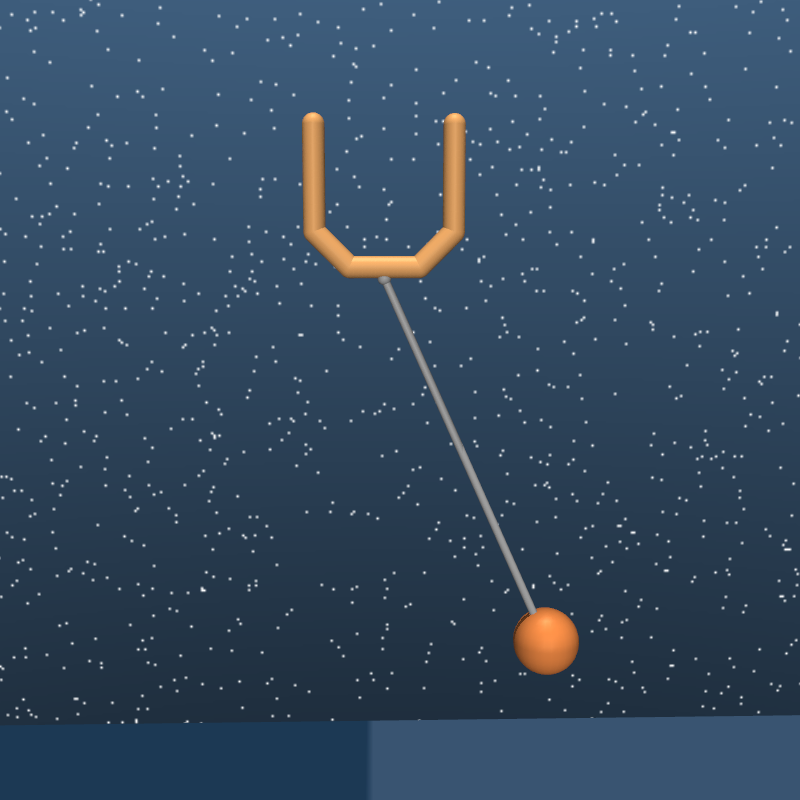}
            \includegraphics[width=0.325\linewidth]{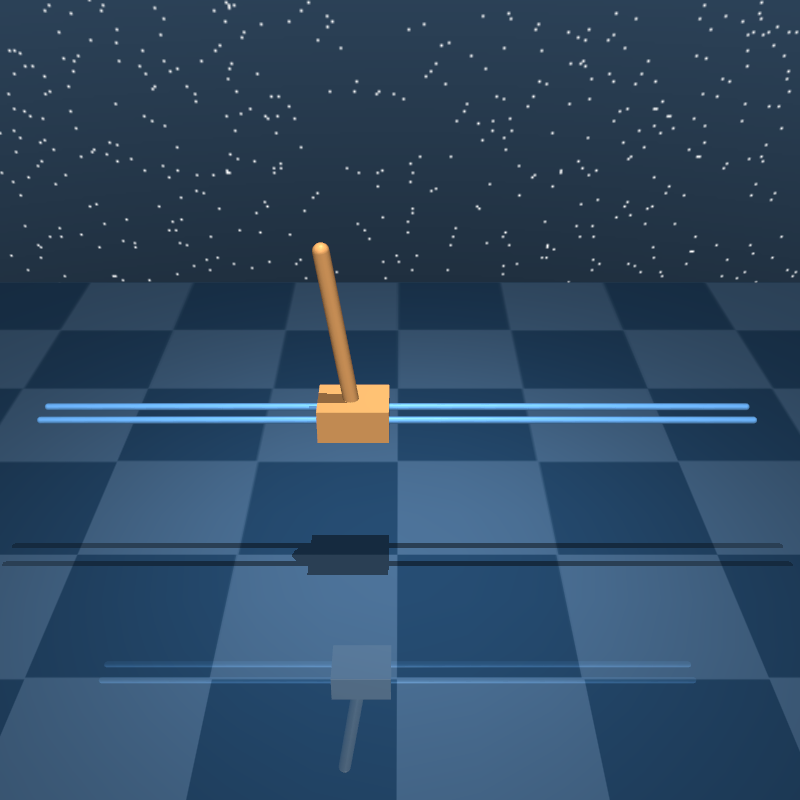}
            \includegraphics[width=0.325\linewidth]{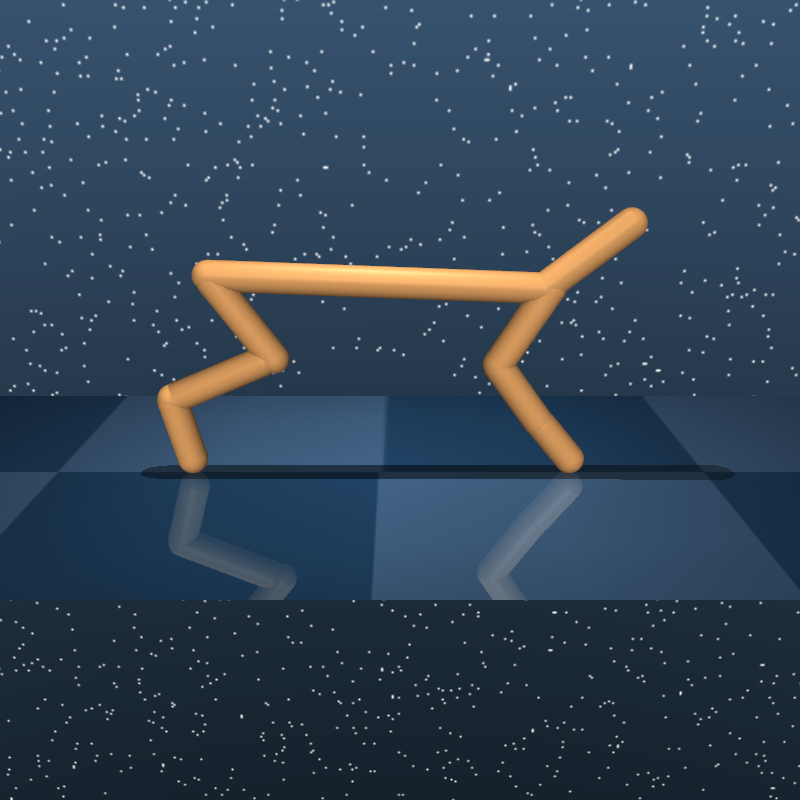}
        \end{minipage}
        \label{fig:dm_control}
        \vspace{-\baselineskip}
        \subcaption{DM Control}
    \end{minipage}
    \begin{minipage}[b]{0.397\linewidth}
        \begin{minipage}[b]{\linewidth}
            \includegraphics[width=0.49\linewidth]{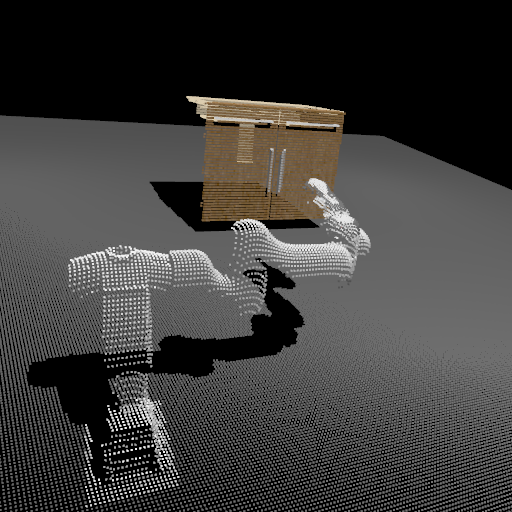}
            \includegraphics[width=0.49\linewidth]{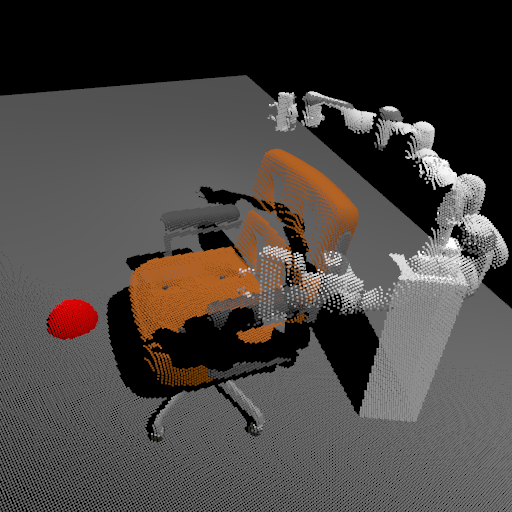}
        \end{minipage}
        \label{fig:maniskill}
        \vspace{-\baselineskip}
        \subcaption{ManiSkill}
    \end{minipage}
    \vspace{-0.2\baselineskip}
    
    \label{fig:env}
    \caption{\textbf{(a)} Sampled environments from the PlaNet Benchmark~\citep{laskin2020reinforcement}: \textit{Ball-in-cup Catch}, \textit{Cartpole Swingup}, \textit{Cheetah Run},
    \textbf{(b)}  Sampled environments from the ManiSkill~\citep{mu2021maniskill} Benchmark: \textit{OpenCabinetDoor} and \textit{PushChair}.
    }
    \label{fig:environments}
  \end{minipage}
\end{figure*}

\begin{wraptable}{r}{6.0cm}
\centering
\scriptsize
\vspace{-1.2em}
\begin{sc}
\setlength{\tabcolsep}{2.0pt}
\begin{tabular}{l|cc}
\toprule
Task & \# Art. pts & \# ground pts \\ %
\midrule
cheetah & 292.9 $\pm$ 2.3 & 944.1 $\pm$ 41.0\\ %
walker & 519.4$\pm$ 29.5& 1691.1$\pm$ 471.2 \\ %
\bottomrule
\end{tabular}
\end{sc}
\caption{Average number of points on the robot articulation and on the ground in the \textit{Cheetah Run} and the \textit{Walker Walk} environments. Point clouds are lifted from $84\times84$ depth images. Since the ground points convey little information, filtering out most of them could accelerate agent training.}
\label{tab:pcd-memory}
\vspace{-1em}
\end{wraptable}

\textbf{Point cloud post-processing.} Given camera matrices, we can obtain raw point cloud observations by back-projecting the RGB-D images. However, the raw observations often contain \textit{redundant} points irrelevant to agent decision-making, e.g. points very far away from the agent. In addition, a large number of points might come from the ground, which conveys little information while consuming a large amount of memory (see Tab.~\ref{tab:pcd-memory} for an illustration). In practice, these points can be filtered out using depth clipping, coordinate clipping, or segmentation mask filtering. After we filter out these irrelevant points, we can then randomly sample a fixed number of remaining points. This allows points to be efficiently batched for agent learning, while allowing agents to focus on the relevant targets and objects critical to decision-making. Later, we will demonstrate that a good point cloud post-processing strategy is crucial for 3D visual RL agents to perform well.

\textbf{Stacking multiple point cloud frames.} For continuous control tasks that depend on inferring velocity-related information across multiple time steps, such as \textit{CheetahRun} in DMC~\citep{tassa2018deepmind}, stacking multiple frames of observations could be beneficial for agent success. For 2D image-based agents, a common practice is to concatenate multiple frames of images along the channel dimension. For 3D point cloud agents, one way to process multiple frames of point clouds is by using a 4D visual backbone such as MinkowskiNet~\citep{Choy20194DSC}. However, these backbones are usually heavy and not computationally efficient for online reinforcement learning. In our work, we adopt a simple and efficient point frame stacking technique, which appends one-hot vectors to point features to indicate which frame each point comes from, and concatenates multiple frames of point clouds along the number-of-points dimension. Agents can then use a lightweight 3D visual backbone, such as PointNet and 3D SparseConvNet, to consume the multi-frame point clouds.

\textbf{Point cloud data augmentations.}
Prior works on 2D image-based RL, such as DrQ~\cite{kostrikov2020image}, have adopted random-shift image augmentation to improve the algorithm's sample efficiency. For 3D point clouds, we can perform similar augmentations by randomly shifting or rotating 3D point coordinates. Details are described below:
\begin{enumerate}
    \item \textit{Random Shift}: Let $O \in \mathbb{R}^{K\times N \times C}$ (see Sec.~\ref{sec:background} for definitions) be a stacked point cloud with $K$ frames. The augmented point cloud $O'$ satisfies $O'[\dots,\textrm{:}3] = O[\dots,\textrm{:}3] + \textrm{broadcast}(\Delta p)$, where $\Delta p \sim \textrm{Unif}(-c, c)^3$ applies a random noise to point cloud coordinates. Note that following prior work~\citep{laskin2020reinforcement,kostrikov2020image}, we apply the same augmentation across multiple frames. We also apply the \textit{same} augmentation to each point, since applying different augmentations could blur the object geometry.
    \item \textit{Random Rotation}: $O'[..., \textrm{:}3] = O[..., \textrm{:}3]R^T$, where $R=\textrm{Euler2Rot}(0,0,\delta)$ and $\delta \sim \textrm{Unif}(-c, c)$. $R$ rotates the point cloud with a random angle along the $z$-axis. One could also use arbitrary rotations in $\mathbb{SO}(3)$, but our preliminary experiments found this harmful.
\end{enumerate}

\section{Experiments}

\begin{figure*}[t!]
    \centering
    \hspace*{-0.08in}\includegraphics[width=0.96\linewidth]{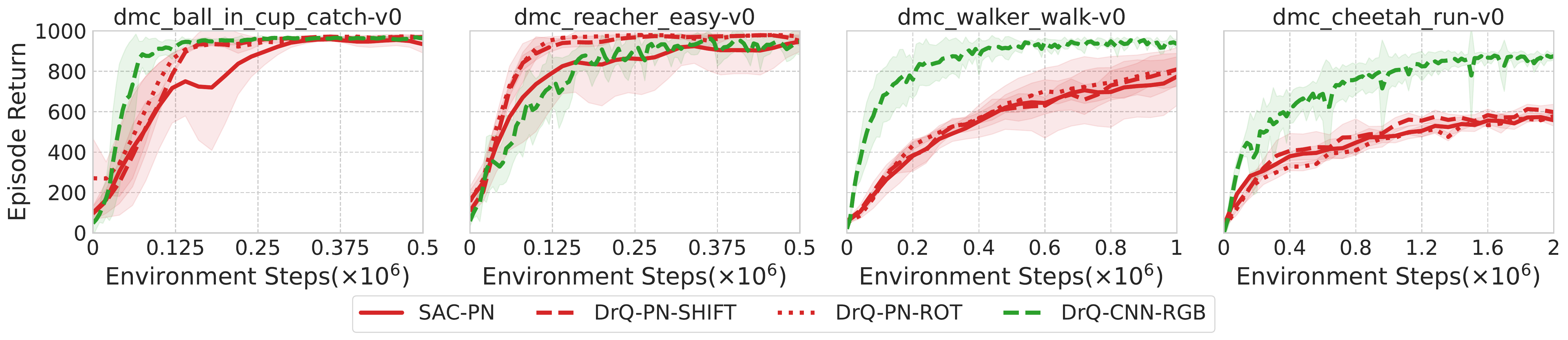}
    \includegraphics[width=\linewidth]{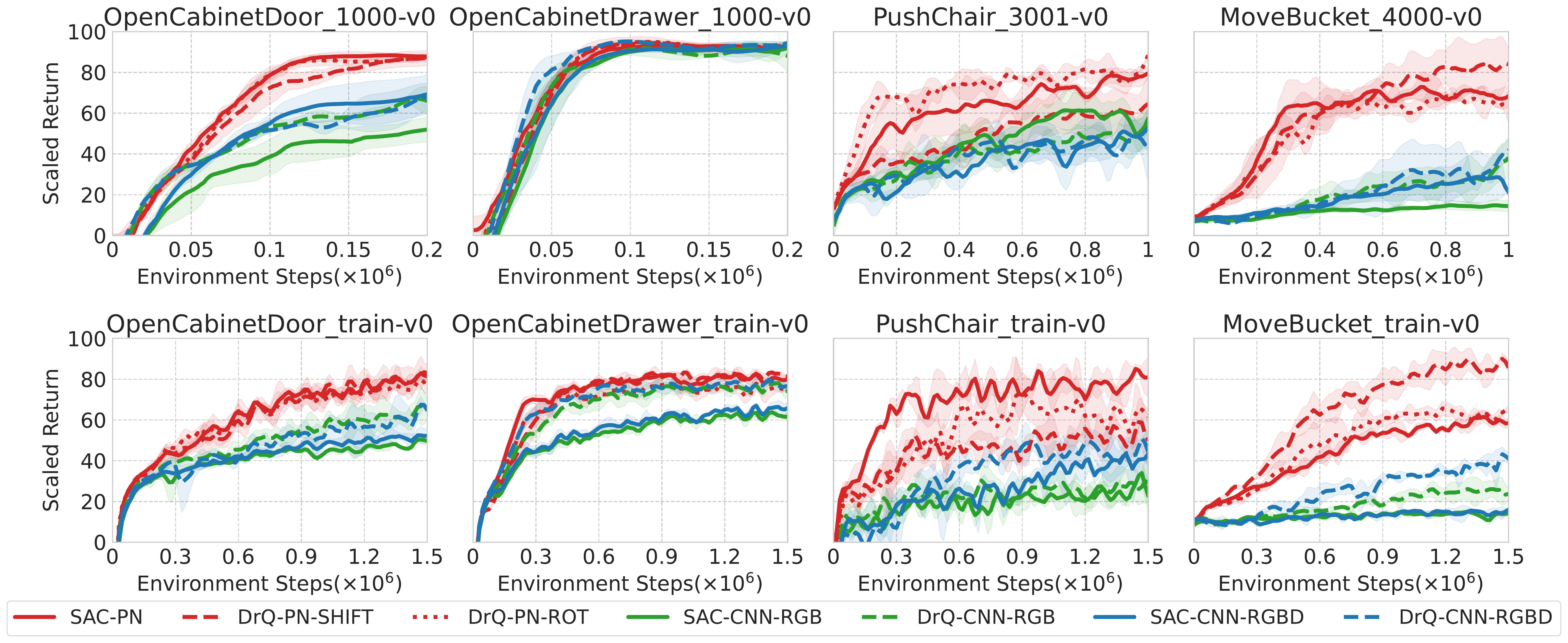}
    \caption{\textbf{(Up)} RGB CNN agents vs. 3D PointNet agents on 4 environments from the PlaNet Benchmark (full results on all 6 environments in Appendix~\ref{app:complete_dmc}). Mean and standard deviation of returns over 3 seeds are shown. All ``SAC'' curves denote agents trained without data augmentations, and all ``DrQ'' curves denote agents with 2D / 3D data augmentations. DrQ-CNN results are obtained from \cite{kostrikov2020image}. \textbf{(Mid, Down)} 3D PointNet agents vs. RGB \& RGBD CNN agents on the ManiSkill Benchmark (both with single object (mid) and with object variations (down); the collections of objects we used for each task are shown in Appendix~\ref{app:obj_variation_maniskill2}.). Point cloud agents have the best sample efficiency and performance on these tasks as they involve heavy 3D spatial relationship reasoning.}
    \label{fig:dmc4_and_maniskill}
\end{figure*}

\begin{figure}[t!]
    \centering
    \includegraphics[width=\linewidth]{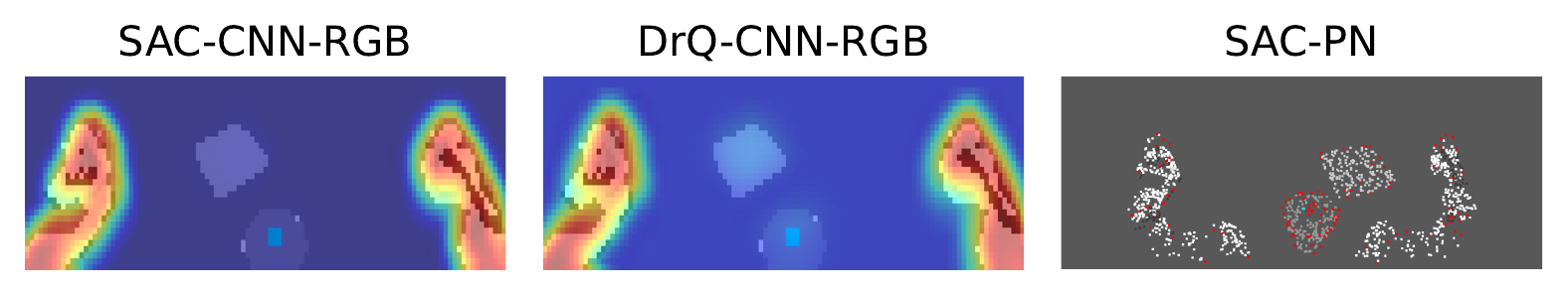}
    \caption{(Left, mid) Activation maps at the final layer of CNN visual backbones for SAC and DrQ agents on the MoveBucket task, where the robot (in egocentric view) moves a bucket containing a blue ball onto the target platform. Red areas denote higher activation, and blue areas denote lower activation. (Right) For PointNet-based SAC agents, red points contribute to at least one dimension of the global visual feature, which is obtained through pointwise feature maxpooling. White points do not contribute to the global visual feature.}
    \label{fig:maniskill_attn_visualization}
\end{figure}

In this section, we compare and analyze the efficacy of 2D image and 3D point cloud-based visual RL across diverse control and robot manipulation tasks. We illustrate some of the environments in Fig.~\ref{fig:environments}. Following prior work~\citep{kostrikov2020image,hafner2019learning,srinivas2020curl}, we adopt 6 control tasks from the widely-used \textbf{DeepMind Control Suite} (DM Control)~\citep{tassa2018deepmind} for image-based RL, known as the \textit{PlaNet Benchmark}. These environments are generally single-agent control tasks, and generally do not involve agent-object or object-object spatial relation understanding. We extend these environments to support point cloud-based learning. In addition, we adopt 4 robotic manipulation tasks from the \textbf{ManiSkill Benchmark}~\citep{mu2021maniskill}, which involve single-arm and dual-arm mobile robots manipulating diverse articulated objects, such as opening cabinet doors / drawers, pushing chairs, and moving buckets. These tasks involve heavy 3D spatial relationship reasoning between agents and objects, such as between the robot hand and the target drawer / door handles in \textit{OpenCabinetDrawer / Door}, and between the two robot hands, the chair, and the goal target in \textit{PushChair}. More details on ManiSkill environments are presented in Appendix~\ref{app:ManiSkill}.

\textbf{Training setup.}
We adopt SAC~\citep{haarnoja2018soft} and DrQ~\citep{kostrikov2020image} as the base RL algorithms. \textit{In all of our training curves, we will use ``SAC'' to denote agents trained without data augmentation, and ``DrQ'' to denote agents trained with 2D or 3D data augmentations}. For RGB and RGB-D-based CNN agents, we followed the implementation, network architectures, and data augmentations of~\cite{kostrikov2020image}. For 3D point clouds agents, we adopt PointNet ~\citep{qi2017pointnet} as the visual backbone, along with the post processing, data augmentation, and frame stacking techniques introduced in Sec.~\ref{sec:design_space}. We will carefully study these design choices in Sec.~\ref{sec:ablations}. More detailed setups are presented in Appendix~\ref{sec:app_more_implementation_details}.

\subsection{Comparing 2D Image vs. 3D Point Cloud RL Across Tasks}
\label{sec:results}

We first compare the performances between 2D RGB/RGBD and 3D point cloud-based visual RL agents on the PlaNet Benchmark and the ManiSkill Benchmark. Results are presented in Fig.~\ref{fig:dmc4_and_maniskill}.  We obtain distinct findings across these two benchmarks. On the PlaNet Benchmark, where the tasks are primarily 2D locomotion tasks (e.g., \textit{Cheetah Run}, \textit{Walker Walk}) that do not involve heavy 3D spatial reasoning, we observe that 2D RGB-based agents tend to outperform 3D point cloud-based agents. On the other hand, on the ManiSkill Benchmark, where the tasks involve heavy agent-object and object-object relationship reasoning in 3D scenes, we find that 3D point cloud agents have the best sample efficiency and performance, which aligns with our motivating example in Sec.~\ref{sec:minimalist}. Interestingly, for a few tasks on the PlaNet Benchmark that involve agent-object relationship reasoning but within a 2D planar space (\textit{Ball in Cup Catch}, \textit{Reacher Easy}), we observe that 3D point cloud agents exhibit similar sample efficiency as 2D agents. Additionally, while we found that data augmentations in 2D images have a significant impact on agent performance across most tasks, data augmentations in 3D point clouds are much less helpful. We will later show that this phenomenon is caused by the inductive bias of the specific 3D backbone architecture we use (i.e. PointNet), which has higher value-estimation robustness than convolution-like architectures.

To gain further insights into policy behaviors, we visualize and compare the activation patterns of final-layer visual features between 2D CNN agents and 3D PointNet agents. We illustrate an example on the MoveBucket task in Fig.~\ref{fig:maniskill_attn_visualization}, where the robot moves a bucket containing a blue ball onto the target platform. We find that for CNN agents, the bucket contributes little to the final visual feature, which is predominantly influenced by robot pixels. On the other hand, for PointNet agents, the final global visual feature incorporates contributions from the robot, the bucket, and the target platform, indicating that it is influenced by the relative relationships between the agent and different objects.

\subsection{Investigating Design Choices and Inductive Biases of 3D Point Cloud RL}
\label{sec:ablations}

As we aim to build and comprehensively evaluate a strong 3D visual RL algorithm, in this section, we conduct a thorough experimental analysis to investigate and ablate the design space we introduced in Sec.~\ref{sec:design_space}. Along this process, we gain deeper insights into the phenomena we discovered in Sec.~\ref{sec:results}.

\begin{figure}[t!]
    \includegraphics[width=\linewidth]{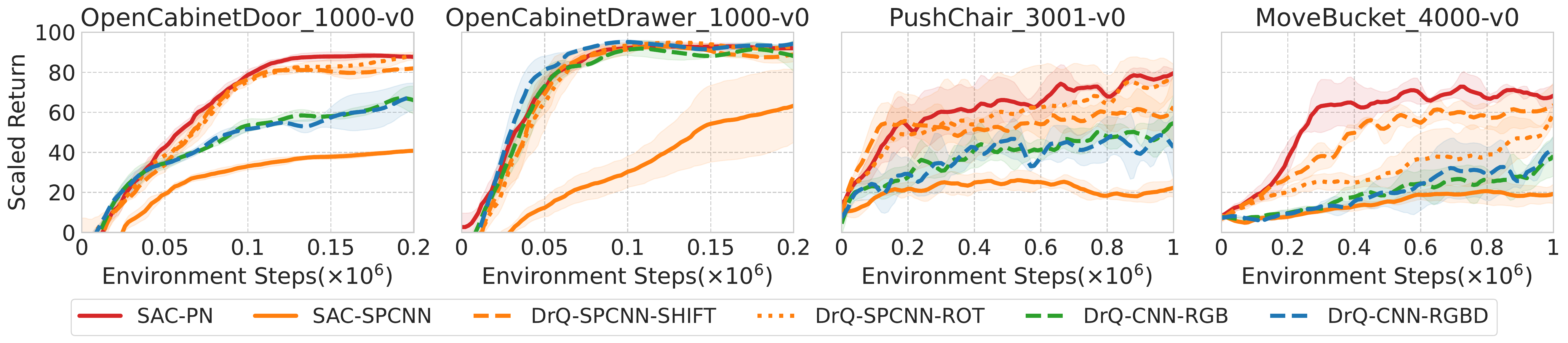}
    \caption{Comparison between 3D point cloud agents using PointNet~\citep{qi2017pointnet} and 3D SparseConvNet~\citep{tang2020searching} as visual backbones. The inductive bias of 3D SparseConvNet is more similar to 2D CNNs.}
    \label{fig:sparseConv}
\end{figure}

\begin{figure}[t!]
    \includegraphics[width=\linewidth]{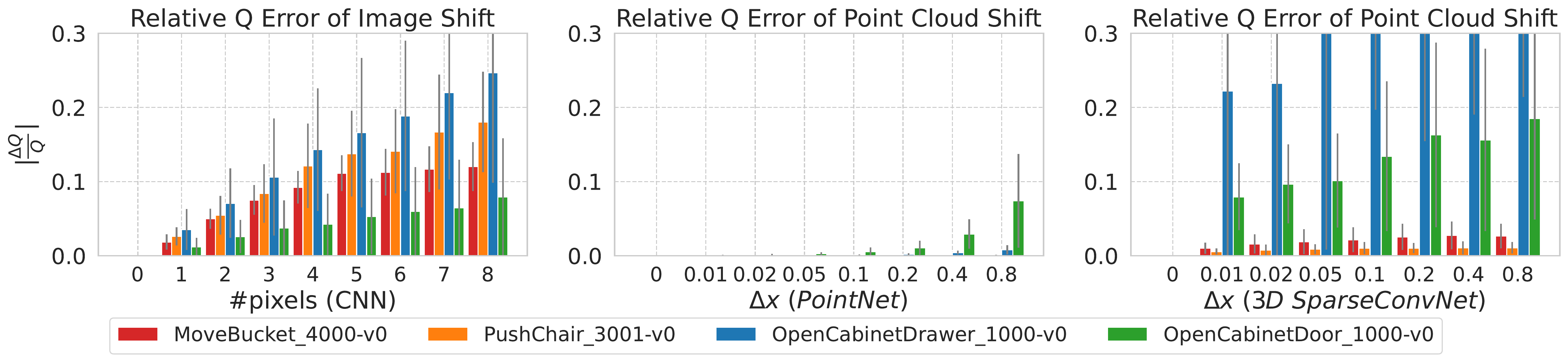}
    \caption{Value network robustness under random shift between a 2D CNN agent, a 3D PointNet agent, and a 3D SparseConvNet agent, measured by the relative $Q$-value changes ($|\frac{\Delta Q}{Q}|$). For point cloud agents, $\Delta x$ indicates a [$\Delta x$, $\Delta x$, $\Delta x$] shift in 3D point cloud positions. A one-pixel random shift in 2D image corresponds to a [0.0272, 0.0276, 0.0003] shift on-average in 3D point positions (thus, for example, 8 pixel shift in images is equivalent to a [0.2176, 0.2208, 0.0024] shift in 3D point positions). The inductive bias of PointNet makes it more robust to visual input perturbations than convolution-based 3D SparseConvNet and 2D CNN.}
    \label{fig:deltaQ}
\end{figure}

\textbf{Performance and representational robustness of different 3D network architectures}: 
In Sec.~\ref{sec:results}, we utilized PointNet as the 3D point cloud backbone, which yields better sample efficiency and performance on tasks that involve substantial 3D spatial relationship reasoning. We also found that random shift and random rotation data augmentations had little impact on PointNet agent performance. To gain further insights into these observations, we seek to answer the following questions: \textbf{(1)} Does the performance advantage hold true for other 3D point cloud backbones? \textbf{(2)} How do these phenomena correlate with the inductive biases of different visual networks?

We first compare the efficacy of PointNet~\citep{qi2017pointnet} vs. 3D SparseConvNet~\citep{tang2020searching}. The latter architecture utilizes sparse convolutions to process point / voxel features, and thus exhibits similar inductive bias as CNNs. Results are shown in Fig.~\ref{fig:sparseConv}. We find that without data augmentations, 3D SparseConvNet performs significantly worse than PointNet. However, data augmentations are particularly beneficial for 3D SparseConvNet, allowing it to perform similarly to PointNet and outperform 2D CNNs. 

Since 3D SparseConvNets behave similarly to CNNs, we aim to explore the specific inductive bias differences between PointNet and these convolution-based architectures that lead to the distinct efficacy of data augmentations. To achieve this, we conduct an experiment that measures the representational robustness between these networks under random perturbations. Specifically, we measure the relative change of the value $Q$ network for a 2D CNN agent, a 3D PointNet agent, and a 3D SparseConvNet agent when there is a random shift in image pixel or point position. Results are presented in Fig.~\ref{fig:deltaQ}. We found that PointNet is exceptionally robust against input perturbations, with little relative change in $Q$-value even with a random shift of 0.2 meters in each dimension of the input point cloud coordinates (which approximately corresponds to 8 pixels of image shift). Intuitively, such representational robustness is especially helpful for learning skills like distance inference that generalizes across scenes, as it prevents agents from memorizing precise locations of objects that appeared during training and helps with solving robotic manipulation tasks. On the other hand, 2D CNN and 3D SparseConvNet are significantly less robust against random shift perturbations, which explains the high efficacy of data augmentations on these architectures. 

\begin{wrapfigure}{r}{0.5\linewidth}
    \vspace{-0.5em}
    \begin{center}
    \includegraphics[width=\linewidth]{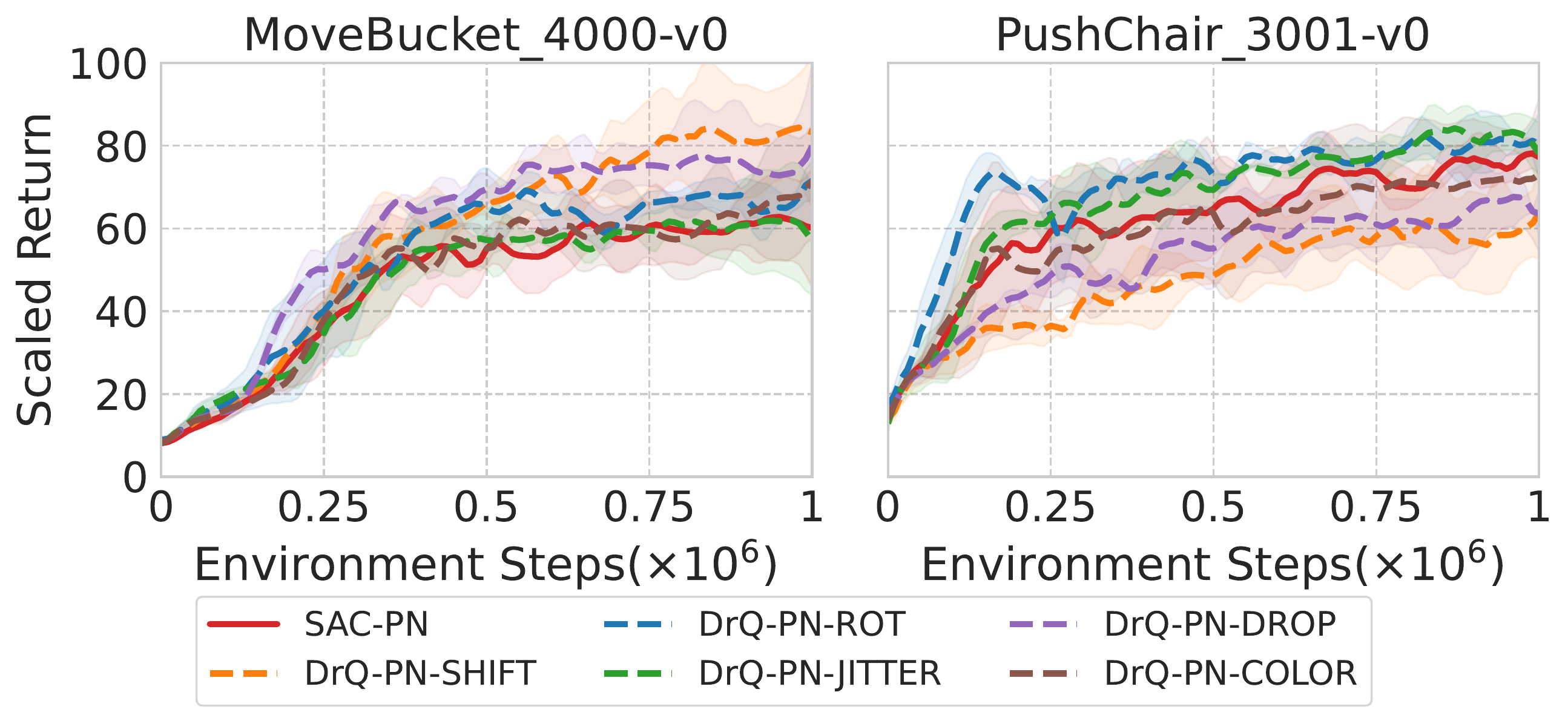}
    \end{center}
    \caption{Efficacy of different 3D point cloud data augmentations for the PointNet agent. None of these augmentations is signficantly beneficial.} 
    \label{fig:moreaugs}
    \vspace{-0.5em}
\end{wrapfigure}

\textbf{Efficacy of More Point Cloud Data Augmentations.} Previously, we examined the efficacy of random shift and random rotation point cloud augmentations as introduced in Sec.~\ref{sec:design_space}. In this paragraph, we further explore and analyze additional point cloud data augmentation strategies: (1) Random Jitter, where we add $\textrm{Unif}(0,0.01)^3$ random noise to point coordinates, and each point receives a different noise; (2) Random Color Shift, where we apply \texttt{torchvision.colorjitter} to the input point cloud with brightness, contrast, saturation, and hue set to 0.5; (3) Random Dropout, which randomly removes $20\%$ of points from the input point cloud. We conduct data augmentation experiments using PointNet agents. Results are shown in Fig.~\ref{fig:moreaugs} (more results in Appendix~\ref{app:complete_augmentations}). We observe that for PointNet agents, none of these augmentations are significantly helpful. This can be attributed to our previous finding that PointNet is already highly robust against input perturbations.

\begin{figure*}[t!]
    \centering
    \includegraphics[width=\linewidth]{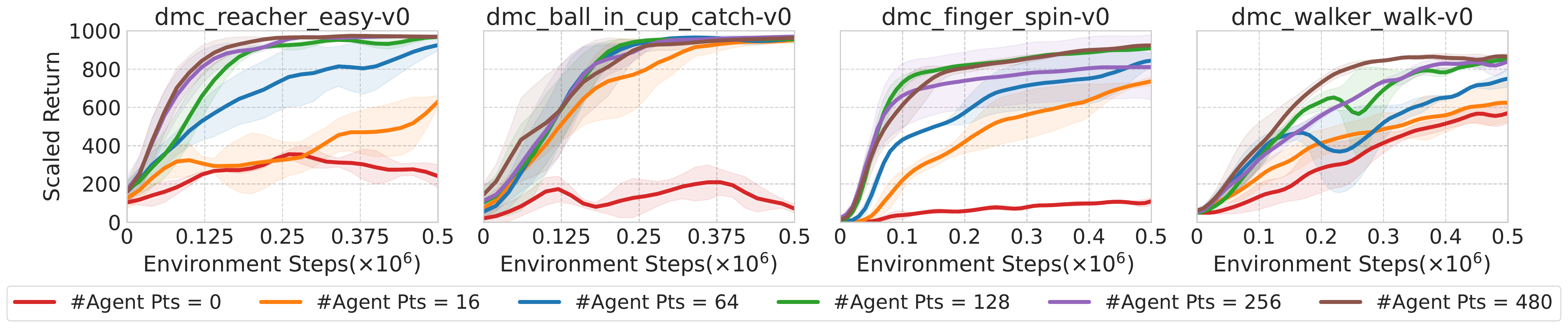}
    \caption{
         Comparison between different point cloud downsampling strategies on the PlaNet Benchmark. We downsample input point clouds to 512 points per frame, and we vary the proportion of points coming from the agent body (the rest of the points are sampled from the ground). 
    }
    \label{fig:downsample}
    \vspace{-0.5em}
\end{figure*}

\begin{figure*}[t!]
    \centering
    \includegraphics[width=\linewidth]{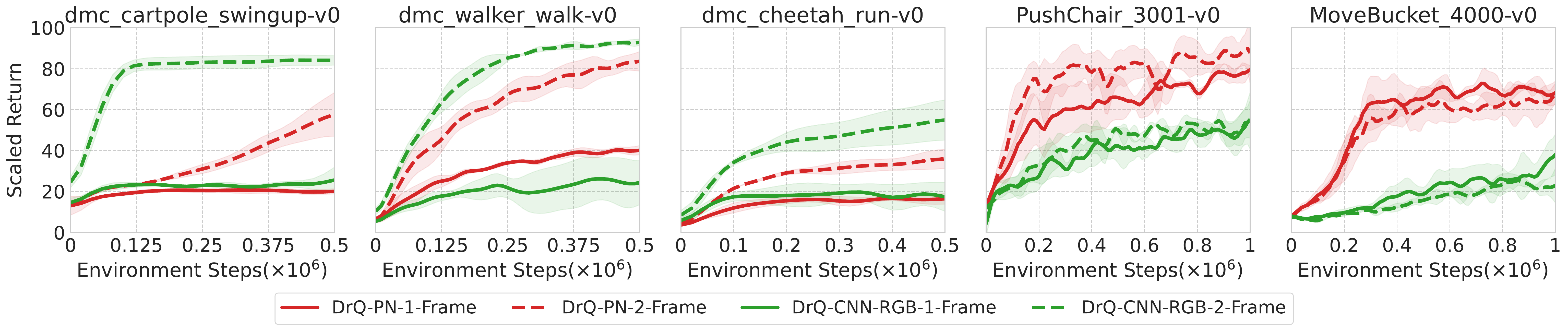}
    \caption{
         Comparison between point cloud agents and image-based agents with single-frame and 2 stacked-frame visual observations.
    }
    \label{fig:multiframe}
    \vspace{-0.5em}
\end{figure*}

\textbf{Impact of point cloud post-processing strategies.}
\label{sec:post-processing}
As previously introduced in Sec.~\ref{sec:design_space}, input point cloud observations often contain redundant points that convey little information for agent decision making. Intuitively, we would like to retain most points critical for agent decision making (e.g., those on the agent body) to ensure effective agent learning, while filtering out most of irrelevant points (e.g., those on the ground) to accelerate agent training. To test our hypothesis, we conduct experiments on the DM Control tasks to explore the impact of different point cloud post-processing strategies. We downsample input point clouds to a fixed number of 512 points per frame, while varying the proportion of points sampled from the agent body and from the ground. Our results in Fig.~\ref{fig:downsample} clearly demonstrates that keeping more points on the agent body leads to better agent performance. This suggests that prioritizing the preservation of points that are crucial to agent decision making can significantly benefit 3D point cloud RL.

\textbf{Efficacy of multi-frame stacking.}
In Sec.~\ref{sec:results}, we observe that 3D point cloud agents are not as effective as image-based agents on some DM Control tasks such as \textit{Cheetah-Run} and \textit{Walker-Walk}. A notable characteristic of these tasks is that for agents to achieve high returns, it is crucial to infer velocity-related dynamics information from multiple observations across time. On the other hand, 3D point cloud agents perform similarly as 2D agents on \textit{Reacher-Easy} and \textit{Ball in cup-Catch} in DM Control, and better than 2D agents on the robot manipulation tasks in \textit{ManiSkill}. For these tasks, a single-frame observation already offers sufficient information to infer the relationship between the agent and the object, making it adequate for successful decision making. We therefore conjecture that our point cloud agents are not yet proficient at inferring agent dynamics information across multiple frames, even with our frame stacking technique introduced in Sec.~\ref{sec:design_space}. As a diagnosis experiment, we compare the performance of point cloud and image-based agents using single-frame and 2 stacked-frame observations. Results are presented in Fig.~\ref{fig:multiframe}. We observe that image and point cloud agents perform similarly when only single-frame observations are provided. However, when agents receive two consecutive frames of observations, the performance boost for point cloud agents is much smaller than the boost for image-based agents on DM Control tasks like \textit{Cheetah-Run} and \textit{Walker-Walk}. This suggests that our current approach of integrating 3D point cloud information across multiple time steps might not be effective as the RGB counterpart, especially on tasks that involve heavy agent dynamics inference across time. We encourage future works to design more effective approaches that can more effectively integrate time information across multiple 3D frames.

\section{Conclusion \& Future Work}
In this work, we conducted the first systematic study on the efficacy of 3D visual RL from point clouds, and compared it with well-established RL from 2D RGB/RGB-D representations. Through a minimalist motivating example as well as a range of control and robotic manipulation tasks, we show that 3D point cloud representations are particularly beneficial on tasks where agent-object/object-object spatial relationship reasoning plays a crucial role, and achieves better sample complexity and performance than 2D image-based agents. Moreover, we carefully investigate the design choices for 3D point cloud RL agents from perspectives such as network inductive bias, representational robustness, data augmentation, and data post-processing. We hope that our study provides insights, inspiration, and guidance for future works on 3D visual RL. A limitation of our work is that we currently focus on online RL with relatively short-horizon tasks. In the future, we would like to extend our work to demonstration-based RL methods along with longer-horizon tasks. In addition, as written in Sec.~\ref{sec:ablations}, we would also like to explore more effective multi-frame point cloud stacking strategies to better benefit tasks that involve agent dynamics inference across time.

\section*{Acknowledgements}

This work is in part supported by Qualcomm AI and AI
Institute for Learning-Enabled Optimization at Scale (TI-
LOS).

\bibliography{main}
\bibliographystyle{plain}

\newpage
\appendix
\onecolumn

\section{More Implementation and Environment Details}
\label{sec:app_more_implementation_details}
For the experiments in Sec.~\ref{sec:results}, we follow the implementation of Soft Actor-Critic(SAC) described in~\citep{haarnoja2018soft}, including clipped Double-Q learning and squashed Gaussian head.
For DrQ~\citep{kostrikov2020image}, we follow the original implementation and use $K=2$, $M=2$, and a random shift of 4 pixels for all RGB and RGB-D images. To augment point cloud observations, we shift point coordinates with a noise of $\textrm{Unif}(0,0.15)$ along all axes or rotate point coordinates around the z-axis with a noise of $\textrm{Unif}(0,0.15)$ radian. The same random shift or random rotation augmentation is applied to all points.

In addition, we follow previous works~\citep{laskin2020reinforcement, kostrikov2020image} and apply the same data augmentation across multiple stacked visual observation frames. For different observations in a single sampled batch, we still apply different data augmentations.

\subsection{DeepMind Control Suite (PlaNet Benchmark)}
\subsubsection{Hyperparameters} 
The following hyperparameters are used to evaluate both CNN and point cloud-based SAC agents on DM Control environments. 
\label{app:DMC}
\begin{table}[h]
\centering
\small
\begin{tabular}{r|cc}
\toprule
Hyperparameters & Value\\
\midrule
Optimizer & Adam \\
Learning rate & $1 \times 10^{-3}$ \\
Learning rate (Entropy $\alpha$) & $1 \times 10^{-3}$ \\
Discount Factor($\gamma$) & 0.99 \\
Initial Temperature & 0.1 \\
Target Update Interval & 2 \\
Actor Update Interval & 2 \\
Actor Log Standard Deviation Bounds & $[-10 ,2]$ \\
Target Soft Update Coefficient (Visual Backbone) & 0.05 \\
Target Soft Update Coefficient & 0.01 \\
Replay Buffer Size & 100000 \\
Warm Up Steps & 1000 \\
Batch Size & 256 \\
\bottomrule
\end{tabular}
\vspace{1em}
\caption{Hyperparameters for SAC on DM Control tasks.}
\label{tab:DMC_param}
\end{table}

\subsubsection{Observation Details}

Following the implementation of \citep{kostrikov2020image}, our RGB and RGB-D image inputs have a resolution of $(84 \times 84)$. For 3D point cloud observations, they are generated by back-projecting RGB-D images of the same resolution using the known camera matrices. We then apply post-processing techniques as described in Section~\ref{sec:design_space} and Appendix~\ref{app:pcd_process}. Point cloud features consist of 3-dimensional $xyz$ coordinates along with 3-dimensional RGB colors.

\subsubsection{Network Architectures}
We use the following visual backbone architectures for the PlaNet Benchmark. The visual backbone is shared between the policy and the value networks.
\begin{itemize}
    \item PointNet: Our PointNet uses 3 layers of MLP with dimensions of $[64, 128, 256]$ before maxpooling over the number-of-points dimension to obtain a global feature. We apply Layer Norm and ReLU activation after every MLP layer except the last layer. After the global feature is obtained, we use a linear layer followed by a Layer Normalization layer to map the output features to 50 dimensions.
    \item CNN: Our CNN visual backbone follows the implementation of DrQ~\cite{kostrikov2020image}. 
    \item Actor-critic Networks: We follow the implementation of actor and critic value networks in \cite{kostrikov2020image}. We set the critic network input dimension to be $50 + action\_shape$ and the actor network input dimension to be 50. We also ensure that visual features extracted from both PointNet and CNN have 50-dimensions. 
\end{itemize}

\subsubsection{Point Cloud Post-Processing}
\label{app:pcd_process}
We employ the following Point cloud post-processing procedure for training point cloud-based agents: 
\begin{enumerate}
    \item We obtain raw point clouds by back-projecting RGB-Depth images with the known camera matrix.
    \item All points whose distances to the camera are further than a threshold are dropped. The default threshold is 10.
    \item We determine whether a point is on the ground by selecting all points whose $z$ coordinate is smaller than the minimum $z$ coordinate of all points plus a small tolerance value (set to 0.008). Points that do not belong to the ground is assumed to belong to the agent body / agent articulation.
    \item Points on the agent articulation and on the ground are uniformly downsampled to a target number of points, or padded if the number of points is not enough. The target numbers of points on the agent articulations for each task are described in Table~\ref{tab:dmc_num_points}. The target number of points on the grounds is set to $\frac{1}{3}$ of the target number of points on the agent articulation.
\end{enumerate}

\begin{table}
    \centering
    \begin{tabular}{c|c}
    \toprule 
    Environment Name & Target Number of Agent Articulation Points \\
    \midrule 
    Cheetah Run & 256 \\
    Walker Walk & 384 \\
    Finger Spin & 384 \\
    Cartpole Balance & 256 \\
    Reacher Easy & 256 \\
    Ball-in-cup Catch & 128 \\
    \bottomrule
    \end{tabular}
    \vspace{1em}
    \caption{Target numbers of points on the agent articulation on each of PlaNet Benchmark tasks}
    \label{tab:dmc_num_points}
\end{table}

\subsection{ManiSkill Benchmark}
\subsubsection{Task Descriptions}
The ManiSkill benchmark consists of the following environments:
\begin{enumerate}
    \item OpenCabinetDoor: A single-arm mobile robot moves to the cabinet and opens a designated cabinet door. A cabinet can contain more than one doors.
    \item OpenCabinetDrawer: A single-arm mobile robot moves to the cabinet and opens a designated cabinet drawer. A cabinet can contain more than one drawers.
    \item PushChair: A dual-arm mobile robot pushes an under-actuated swivel chair to the target location marked by a red sphere on the ground.
    \item MoveBucket: A dual-arm mobile robot picks up the bucket on the ground and moves it onto the target platform. A ball is contained in the bucket, and agents should ensure that the ball stays in the bucket throughout a trajectory.
\end{enumerate}
Details of the observation space and action space are described in the original paper~\citep{mu2021maniskill}.

\subsubsection{Hyperparameters}
We use the following hyperparameters for 2D CNN and 3D point cloud-based SAC agents. For CNN agents, we follow the implementation of DrQ~\cite{kostrikov2020image}. 
\label{app:ManiSkill}
\begin{table}[h]
\centering
\small
\begin{tabular}{r|cc}
\toprule
Hyperparameters & Value\\
\midrule
Optimizer & Adam \\
Learning rate & $1 \times 10^{-3}$ \\
Learning rate (Entropy $\alpha$) & $1 \times 10^{-3}$ \\
Discount Factor($\gamma$) & 0.95 \\
Initial Temperature & 0.1 \\
Target Update Interval & 8 \\
Actor Update Interval & 8 \\
Actor Log Standard Deviation Bounds & $[-10, 2]$ \\
Target Soft Update Coefficient (Visual Backbone) & 0.05 \\
Target Soft Update Coefficient & 0.01 \\
Replay Buffer Size & 100000 \\
Warm Up Steps & 1000 \\
Batch Size & 256 \\
\bottomrule
\end{tabular}
\vspace{1em}
\caption{Hyperparameters for SAC on ManiSkill tasks.}

\label{tab:ManiSkill_param}
\end{table}

\subsubsection{Observation Details}
Our RGB and RGB-D observations have resolutions of $(125 \times 50)$. Point cloud observations are natively supported by the ManiSkill environments, and we downsample input point clouds to 1200 points using the strategy described in~\citep{mu2021maniskill}. The environment also provides segmentation masks, so we append them to the pixel / point features.

\subsubsection{Network Architectures}
The network architectures we used in ManiSkill tasks are mostly the same as those in DM Control Tasks, with the following changes made:
\begin{enumerate}
    \item We changed the dimensions of the 3 shared MLP layers of the PointNet to $[128,128,256]$.
    \item We increased the output dimension of both CNN and PointNet encoders to 128.
    \item We also used SparseConvNet in our study on ManiSkill environments. The SparsesConvNet contains 3 3d convolution layers with feature map dimensions $[64,128,128]$. The kernel sizes are all $(3 \times 3 \times 3)$ and the strides are all 1. Layer Norm is applied after each layer of 3D convolution, and ReLU is used as the activation function except the last layer. To convert points to voxels inputs for SparseConvNet, we first use a 2-layer MLP to project point features to 128 dimensions, then voxelize these features following the implementation of SparseConv.
\end{enumerate}
In addition, following~\citep{mu2021maniskill}, we concatenate encoded visual features with robot proprioceptive states before feeding them into the actor and critic networks. Thus, the actor network has an input dimension of 128 + proprioceptive\_state\_dim, and the critic network has an input dimension of 128 + proprioceptive\_state\_dim + action\_shape. 
\subsubsection{Pointcloud post-processing}
We follow the original ManiSkill paper~\citep{mu2021maniskill} to post-process point clouds. Specifically, we remove all points on the ground, and keep a target number of 800 points on the target object and 400 points on the robot. Dummy points are padded if the number of points is not enough.

\subsection{Motivating Example: 3D Reacher}
\label{app:motivating_example}
\subsubsection{Hyperparameters}
The hyperparameters for image-based and point cloud-based SAC agents are shown in Table \ref{tab:motivating_param}.
\begin{table}[h!]
\centering
\small
\begin{tabular}{r|cc}
\toprule
Hyperparameters & Value\\
\midrule
Optimizer & Adam \\
Learning rate & $1 \times 10^{-3}$ \\
Learning rate (Entropy $\alpha$) & $1 \times 10^{-3}$ \\
Discount Factor($\gamma$) & 0.99 \\
Initial Temperature & 0.1 \\
Target Update Interval & 2 \\
Actor Update Interval & 2 \\
Actor Log Standard Deviation Bounds & $[-10 ,2]$ \\
Target Soft Update Coefficient (Visual Backbone) & 0.05 \\
Target Soft Update Coefficient & 0.01 \\
Replay Buffer Size & 10000 \\
Warm Up Steps & 300 \\
Batch Size & 128 \\
\bottomrule
\end{tabular}
\vspace{1em}
\caption{Hyperparameters for SAC on the motivating example.}
\label{tab:motivating_param}
\end{table}

\subsubsection{Architectures}
The network architectures used for our motivating example are similar to those used for the DM Control tasks, except for the following changes:
\begin{enumerate}
    \item We reduce the dimensions of the 3 shared MLP layers of the PointNet to $[32,64,128]$ for computational efficiency.
    \item We set the output dimension of both CNN and PointNet encoders to 128.
\end{enumerate}

We employ the following Point cloud post-processing procedure for training point cloud-based agents: 
\begin{enumerate}
    \item All points with a depth larger than 6 meters are dropped from the observations
    \item Each point cloud observation is uniformly downsampled to contain 128 total pints. 
\end{enumerate}

\newpage
\section{Complete Results for PlaNet Benchmark}
\label{app:complete_dmc}
In Sec.~\ref{sec:results} and Fig.~\ref{fig:dmc4_and_maniskill}, we presented results on 4 of the 6 PlaNet Benchmark tasks. In this section, we present the full results on all 6 PlaNet Benchmark tasks. Curves of DrQ and DrQ-v2 are from their original repos. We observe that point cloud agents have similar performance as image-based agents on tasks that involve planar agent-object relationship reasoning (\textit{Ball in Cup Catch}, \textit{Reacher Easy}), and has worse performance on tasks where agent dynamics inference across time plays a crucial role in achieving high returns (\textit{Finger Spin}, \textit{Walker Walk}, \textit{Cheetah Run}).

\begin{figure*}[h]
    \centering
    \includegraphics[width=\linewidth]{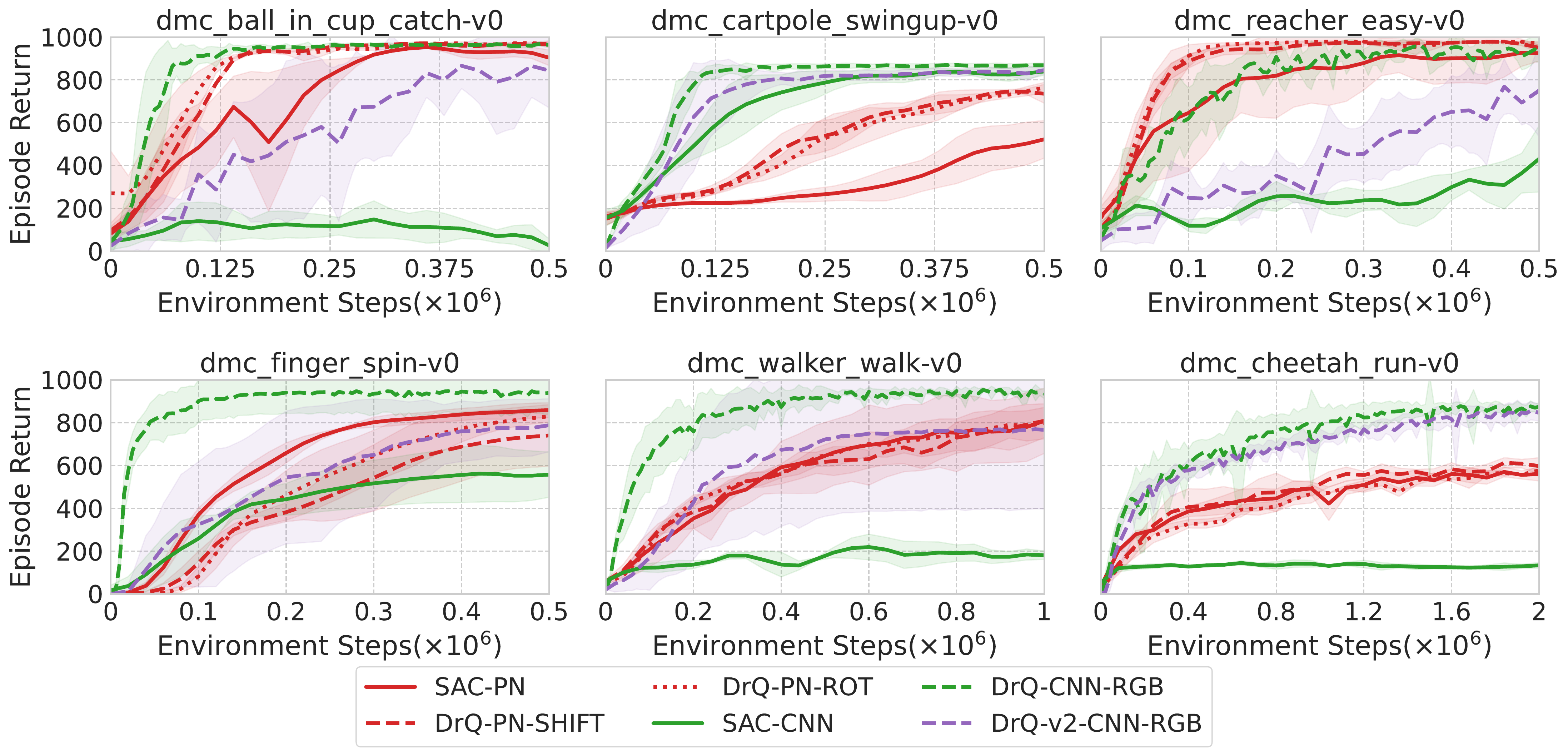}
    \caption{Full results on all 6 PlaNet Benchmark tasks.}
    \label{fig:dmc_all}
\end{figure*}

\section{Complete Results of PointNet Data Augmentations}
\label{app:complete_augmentations}
\begin{figure*}[h]
    \centering
    \includegraphics[width=\linewidth]{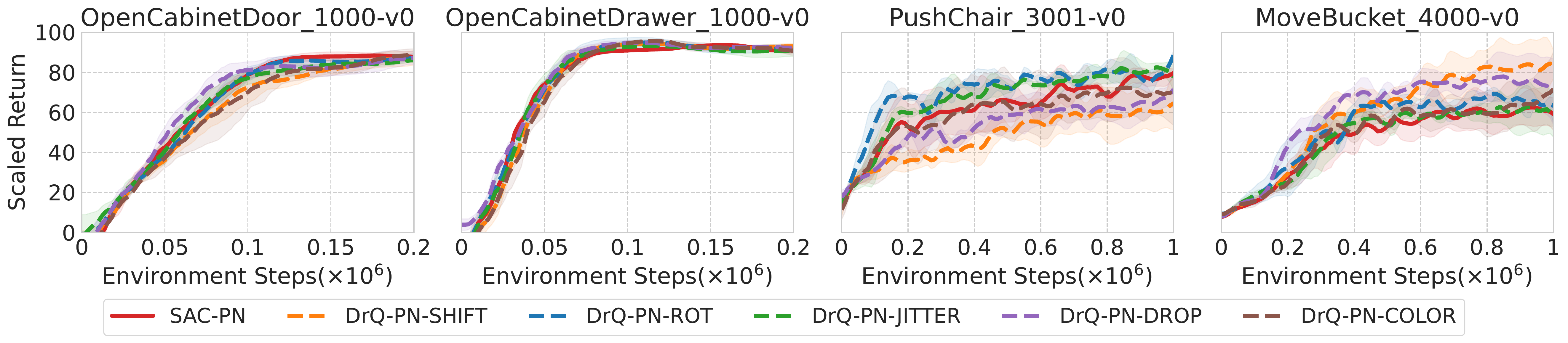}
    \caption{Full results on different point cloud data augmentations for PointNet agents on the ManiSkill Benchmark.}
    \label{fig:augmentations_all}
\end{figure*}

In Section \ref{sec:ablations}, we provided partial results on various point cloud data augmentations and their impact on performance, utilizing PointNet as the visual backbone. In this section, we complement those results by presenting complete findings in Fig.~\ref{fig:augmentations_all}. We observe that none of point cloud data augmentations are significantly helpful for PointNet, which can be attributed to our findings in Sec.~\ref{sec:ablations} that PointNet is already highly robust to input perturbations.

\newpage
\section{ManiSkill Benchmark with Object Variations}
\label{app:obj_variation_maniskill2}
The object ids we use for each task is:

\textbf{OpenCabinetDrawer:} 1000, 1004, 1005, 1013, 1016, 1021, 1024, 1032, 1035, 1038, 1040, 1044, 1045, 1052, 1054, 1061, 1066, 1067, 
1079, 1082

\textbf{OpenCabinetDoor:} 1000, 1001, 1002, 1006, 1007, 1017, 1018, 1025, 1026, 1027, 1028, 1031, 1034, 1038, 1039, 1041, 1042, 1044, 
1045, 1046, 1047, 1049, 1051, 1052, 1054, 1057, 1060, 1061, 1062, 1063, 1064, 1065, 1067, 1068, 1073, 1075, 1077, 1081

\textbf{PushChair:} 3001, 3003, 3005, 3008, 3010, 3013, 3016, 3020, 3021, 3022, 3024, 3025, 3027, 3030, 3031, 3032, 3038, 3045, 3050, 3051, 3063, 3070, 3071, 3073, 3076

\textbf{MoveBucket:} 4000, 4003, 4006, 4008, 4009, 4010, 4012, 4016, 4017, 4018, 4019, 4020, 4021, 4022, 4023, 4024, 4025, 4032, 4035, 4043, 4044, 4051, 4052, 4055, 4056

\end{document}